\pdfoutput=1

\documentclass[11pt]{article}

\usepackage[review]{ACL2023}
\usepackage{booktabs}

\usepackage{times}
\usepackage{latexsym}
\usepackage{multirow}
\usepackage{graphicx}
\usepackage[T1]{fontenc}

\usepackage[utf8]{inputenc}

\usepackage{microtype}

\usepackage{inconsolata}

\newcommand\citeapos[1]{\citeauthor{#1}'s (\citeyear{#1})}

\usepackage{cleveref}
\usepackage{enumitem}
\usepackage{tipa}
\usepackage{footnote}

\title{More than Just Statistical Recurrence: Human and Machine Unsupervised Learning of M\={a}ori Word Segmentation across Morphological Processes
}

\author{Ashvini Varatharaj \\
  Department of Linguistics, \\
  University of California Santa Barbara \\
  \texttt{ashvinivaratharaj@ucsb.edu} \\\And
  Simon Todd \\
  Department of Linguistics, \\
  University of California Santa Barbara \\
  \& NZILBB, University of Canterbury \\
  \texttt{sjtodd@ucsb.edu} \\}

\begin{document}
\nolinenumbers
{\makeatletter\acl@finalcopytrue
  \maketitle
}

\begin{abstract}
Non-M\={a}ori-speaking New Zealanders (NMS)
are able to segment M\={a}ori words 
in a highly similar way to fluent speakers \citep{panther_morphological_2023}. This ability is assumed to derive through the identification and extraction of statistically recurrent forms. 
We examine this assumption by asking how NMS segmentations compare to those produced by Morfessor, an unsupervised machine learning model 
that operates based on statistical recurrence, across words formed by a variety of morphological processes. Both NMS and Morfessor succeed in segmenting words formed by concatenative processes (compounding and affixation without allomorphy), but NMS also succeed for words that invoke templates (reduplication and allomorphy) and other cues to morphological structure, implying that their learning process is sensitive to more than just statistical recurrence.


\end{abstract}

\section{Introduction}

%

Humans have a powerful ability to build implicit linguistic knowledge incidentally, 
based on passive processes that identify and extract statistically recurrent patterns \citep{saffran_statistical_1996,frank_learning_2013,aslin2017statistical}. For example, New Zealanders who are regularly ambiently exposed to M\={a}ori, but do not speak it, nevertheless have implicit lexical and phonotactic knowledge of M\={a}ori \citep{oh_non-maori-speaking_2020,panther_proto-lexicon_2023} and are able to morphologically segment M\={a}ori words at above-chance levels \citep{panther_morphological_2023}. These findings imply that regular ambient exposure to a language yields a \textit{proto-lexicon}: an implicit memory-store of forms that recur with statistical regularity in the language, including both words and word-parts \citep{ngon_nonwords_2013,johnson_constructing_2016}.

In this paper, we are concerned with the way that the proto-lexicon is constructed, and the way that its construction interacts with language structure. We examine the extent to which the ability of non-M\={a}ori-speaking New Zealanders (NMS) to morphologically segment M\={a}ori words is explained by naive statistical learning, in which their proto-lexicon is assumed to be formed purely through the identification and extraction of statistically recurrent forms in ambient M\={a}ori. To do so, we generate expectations for what morphological segmentation would look like through naive statistical learning processes from Morfessor \citep{creutz_unsupervised_2007,virpioja_morfessor_2013}, an unsupervised Bayesian segmentation model. We compare the segmentations produced by Morfessor to those produced by NMS and examine how they vary across words formed by different morphological processes.

Through two analyses, we argue that NMS do more than a naive statistical learning model would suggest. First, we compare the segmentations of Morfessor and NMS across M\={a}ori words formed by affixation and compounding, both concatenative processes, and words formed by reduplication, a templatic process. We find that both are accurate on words formed by affixation and compounding, but NMS are more accurate on words formed by reduplication, suggesting that NMS identify and extract both statistically recurrent forms and higher-level abstract templates. Then, zooming in on words formed by concatenative processes, we ask whether there are other cues to morphological structure that NMS may be picking up on, by comparing the performance of Morfessor across real Māori words that may contain such cues and constructed words that have the same statistical properties but do not contain any such cues. We find that Morfessor is worse at segmenting real words, suggesting that successful learning by NMS requires sensitivity to more cues than just statistical recurrence.

\section{Background}
\label{sec:background}

\subsection{Statistical learning of language}
\label{sec:background-learning}

How humans learn to extract knowledge from their environment is one of the fundamental questions in cognitive science. Implicit learning -- the process of learning without intention, and even without the awareness of what has been learned \citep{williams_neuroscience_2020} -- is one of the main ways we learn from our surroundings. Implicit learning underlies various essential skills such as language comprehension and production, intuitive decision making, and social interaction \citep{rebuschat_implicit_2015}. A particularly prominent form of implicit learning is \textit{statistical learning}\footnote{While early literature on statistical learning focused narrowly on phonotactic transition probabilities, in this work we use the term more broadly to refer to the learning of any statistical properties of language.}. Statistical learning refers to the process of extracting statistical regularities from input and adapting to them, based on considerations of frequency, variability, distribution, and co-occurrence \citep{saffran_statistical_1996}. Humans are highly sensitive to such statistical regularities and implicitly learn them from birth \citep{bulf_visual_2011,gervain_neonate_2008,teinonen_statistical_2009}. 

While most work on statistical learning has focused on studying infants \citep{saffran_words_2001,pelucchi_statistical_2009} in lab-based setups, recent works have shown that adults are also capable of statistical learning of implicit linguistic knowledge through everyday exposure to a language they don't speak. Non-speakers of M\={a}ori in New Zealand \citep{oh_non-maori-speaking_2020,panther_proto-lexicon_2023} and Spanish in California and Texas \citep{todd_language_2023} show evidence of implicit phonotactic and lexical knowledge of their respective ambient languages. However, this knowledge appears to be weaker in the case of Spanish than in M\={a}ori, and it has been argued that this difference may partly derive from differences in morphological structure \citep{todd_language_2023}.

In addition to implicit phonotactic and lexical knowledge of M\={a}ori, non-M\={a}ori-speaking New Zealanders (NMS) also have the ability to morphologically segment M\={a}ori words in a highly similar way to fluent speakers \citep{panther_morphological_2023}. This ability is facilitated by their possession of a \textit{proto-lexicon} \citep{johnson_constructing_2016,ngon_nonwords_2013}, a large implicit memory-store of the forms of words and word-parts that recur with statistical regularity in the language, called \textit{morphs}. 



\subsection{Morphological segmentation}
\label{sec:background-segmentation}

Unsupervised morphological segmentation provides us an avenue to simulate implicit statistical learning processes. There are many unsupervised morphological segmentation models \citep[e.g.][]{Goldsmith2001, Johnson2007, Eskander2016, Godard2018, Xu2018, Xu2020}. In this work, we use Morfessor Baseline \citep{creutz_unsupervised_2007,virpioja_morfessor_2013}, one of the popular unsupervised morphological segmentation models in the field. Its underlying generative process is highly compatible with a naive model of statistical learning of morphological structure. It assumes that words are formed through the concatenation of morphs, without phonological alternations. It uses a Minimum Description Length framework \citep{rissanen_modeling_1978} to identify the simplest set of morphs that can generate the training data with the highest probability.

Morfessor's statistical learning approach mirrors that which has been assumed for NMS \citet{oh_non-maori-speaking_2020}. Both are learning to segment based on statistical patterns in the language they are exposed to, without getting feedback. In both cases, the learners are identifying recurring forms and extracting them as morphs in a proto-lexicon. By using Morfessor as a baseline of comparison for NMS, we can understand how much of NMS' implicit knowledge is due to simple statistical learning processes.
 

\subsection{The Māori Language}
\label{sec:background-maori}

The Māori language consists of ten consonants <p, t, k, m, n, ng, w, r, wh, h>, five short vowels <a, e, i, o, u>, and five long vowels <ā, ē, ī, ō, ū>. The orthographic system is highly transparent: each grapheme or digraph corresponds to a unique phoneme. The basic timing unit is the mora, where short vowels count as one mora each and long vowels count as two \citep{Harlow2007}. The syllable structure is (C)V(V), but is often treated as (C)V for modeling purposes because of the complexity of distinguishing diphthongs from sequences of monopthongs \citep{Bauer1993,oh_non-maori-speaking_2020}. There is a general minimality constraint which states that (content) words and morphs consist of at least two moras \citep[p.\ 544]{Bauer1993}, and it has been argued that words consisting of four or more moras are highly likely to be morphologically complex \citep{Krupa1968,DeLacy2003}.



There are three main morphological processes in M\={a}ori: reduplication, affixation, and compounding \citep{Bauer1993,Harlow2007}. Reduplication consists of the repetition of part of a base, following one of many templates \citep[see e.g.][]{Keegan1996,todd_unsupervised_2022}. Affixation and compounding both consist of the concatenation of morphs, but in the affixation case the same morphs recur across many words, whereas in the compounding case the morphs do not recur as widely. 

There are no morphophonological alternations in M\={a}ori, but there is affix allomorphy, in which affixes follow phonological templates, with different thematic consonants that are to some extent predictable \citep{ParkerJones2008}. This allomorphy is restricted to the passive and nominalizing suffixes, each of which has default and non-default allomorphs that are or are not consistent with major phonological templates (passive: \textit{-Cia}; nominal: \textit{-Canga}; both for thematic consonant \textit{C}). 

The three morphological processes in M\={a}ori are consistent with Morfessor's assumptions to different extents: reduplication is not; affixation is, but allomorphic templates are not; and compounding is. This allows us to examine how the degree to which Morfessor reflects NMS morphological segmentations is affected by morphological structure.



\section{Analysis 1: Sensitivity to templates}
\label{sec:analyis1}

Our first analysis examines Morfessor and NMS segmentations of Māori words formed through different morphological processes. For each learner, we identify the sensitivity to general templates and the importance of morphological concatenativity by comparing segmentation performance across words formed by reduplication and words formed by affixation or compounding (\Cref{sec:analyis1-a}), as well across cases of affixation that follow salient allomorphic templates to different extents (\Cref{sec:analyis1-b}). This analysis reveals how unsupervised learning of morphological segmentation is sensitive to linguistic structure, and the extent to which the underlying assumptions of Morfessor make it a plausible model of naive statistical learning of morphological segmentation in humans.



\subsection{Data}
\label{sec:analysis1-data}

The analysis is conducted over a subset of words that \citet{panther_morphological_2023} grouped according to morphological and phonological criteria, which we aggregated into categories based on the morphological processes they likely represent. We filtered each category to only include words that were segmented by a fluent M\={a}ori speaker (MS) in a way that is consistent with the assumed morphological process, based on segmentations collected by \citet{oh_non-maori-speaking_2020}. After this filtering, the analysis is based on 3,919 words, categorized as follows:

\begin{description}[itemsep=3pt, parsep=0pt, topsep=6pt, labelindent=0.5em, leftmargin=1em]
\item[Monomorphemic:] Words consisting of 2 or 3 moras ($N=622$ / $295$, respectively) that did not receive any boundaries in the MS segmentation.
\item[Reduplication:] Words that were segmented by the MS in a manner consistent with one of four reduplication templates\footnote{The reduplication category includes some cases where there is both reduplication and compounding. We assess the placement of all boundaries in such cases, regardless of whether they separate the reduplicant from the base or one compound component from another.} \citep[see][for details]{panther_morphological_2023}: total ($N=439$), right ($N=276$), left ($N=111$), or left with lengthening ($N=36$). 
Total reduplication is the most salient of these templates.
\item[Affixation:] Words in which the MS recognized either the 
causative prefix \textit{whaka-} ($N=296$), a passive suffix ($N=437$), or a nominalizing suffix ($N=203$). 
The suffixes have many allomorphs which differ in terms of frequency and consistency with a major phonological template.
\item[Compounding:] Words that consist of four or more moras, without reduplication or affixation, and for which the MS identified at least one boundary ($N=1204$; a subset of the `polymoraics' explored by \citealp{panther_morphological_2023}). 
\end{description}

For each word, we compare a gold standard segmentation provided by the MS to the segmentations provided by Morfessor and NMS.  The Morfessor segmentations were obtained from a model trained with default settings (using the implementation of \citealp{virpioja_morfessor_2013}) on 19,595 word types from the Te Aka dictionary \citep{Moorfield2011}. The NMS segmentations are based on data collected by \citet{panther_morphological_2023} in a word-splitting task, where NMS participants split orthographically-presented words into pieces by placing any number of boundaries at any site between two letters. To aggregate segmentations of a single word across participants, we used a majority-vote approach: we coded each site as containing a boundary if and only if the majority of participants who responded to that word placed a boundary there.

\subsection{Analysis 1A: Morphological processes}
\label{sec:analyis1-a}

We first analyze the degree to which segmentations by Morfessor and NMS match the gold standard segmentations, across categories of words formed by different morphological processes. We examine variation across categories, as well as how this variation differs between learners.

\subsubsection{Methods}
\label{sec:analysis1-a-methods}

There are many metrics that compare a learner's morphological segmentations to a gold standard \citep{virpioja-etal-2011-empirical}. We use the simple metric of boundary precision and recall, which considers each potential boundary site independently. Precision in this context refers to the proportion of the sites identified by the learner as containing a boundary that also contain a boundary in the gold standard segmentation. Recall refers to the proportion of the sites containing a boundary in the gold standard segmentation that are identified by the learner as containing a boundary. We take a macro-averaging approach: we calculate precision and recall separately for each word, then average each metric across all words in each category. If precision and recall are both undefined for a word (i.e., if the gold standard segmentation contains no boundaries and the learner does not identify any), we set them both to 1; if only one metric is undefined, we set that metric to 0.

\subsubsection{Results}
\label{sec:analysis1-a-results}

The macro-averaged precision and recall for Morfessor and NMS across the four categories of words are shown in \Cref{tab:analysis1-results2}. 

\begin{table}
\centering
\caption{Macro-averaged precision and recall for Morfessor and NMS across categories of words formed by different morphological processes.}
\label{tab:analysis1-results2}
\begin{tabular}{@{}lcccc@{}}
\toprule
& \multicolumn{2}{c}{Morfessor} & \multicolumn{2}{c}{NMS} \\ 
\cmidrule(lr){2-3} \cmidrule(lr){4-5}
Category & Prec. & Rec. & Prec. & Rec. \\ 
\midrule
monomorphemic & 0.66 & 0.66 & 0.79 & 0.79 \\
reduplication & 0.58 & 0.51 & 0.85 & 0.86 \\
affixation & 0.92 & 0.90 & 0.70 & 0.70 \\
compounding & 0.88 & 0.91 & 0.84 & 0.84 \\
\bottomrule
\end{tabular}
\end{table}

For monomorphemic words, both learners show indications of oversegmentation, via low precision and recall that result from placing boundaries where they shouldn't exist. NMS appear to show less oversegmentation than Morfessor, suggesting that they may be more sensitive to word minimality constraints based on moraic weight \citep[p.\ 544]{Bauer1993}. This tendency toward oversegmentation does not stand out for either learner across other categories: precision and recall are fairly balanced for both learners across all categories, indicating a general balance between oversegmentation and undersegmentation.


For words formed by reduplication, a templatic process, NMS show better performance than Morfessor. This difference is made even clearer when considering performance on reduplication in relation to affixation and compounding (concatenative processes): for Morfessor, performance on reduplication is notably worse than performance on affixation and compounding, but for NMS, it is not. This result suggests that NMS may be sensitive to abstract reduplication templates that Morfessor cannot capture \citep{todd_unsupervised_2022}, and thus that their recognition of such templates may boost implicit learning above and beyond that expected from simple statistical learning of recurrent forms. In support of this suggestion, we found that Morfessor has worst performance on the subset of words formed by total reduplication, the most salient reduplication template, whereas NMS has best performance on this subset (precision/recall for Morfessor: 0.35/0.36; for NMS: 0.95/0.97).

For words formed by affixation and compounding, both concatenative processes, Morfessor performs well, suggesting that such words facilitate implicit learning of morphs via naive statistical learning. Nevertheless, it is somewhat surprising that Morfessor did not perform even better for these words, given that they exactly match the assumptions of its underlying generative model. This suggests that the morphological structure of M\={a}ori, as captured by the gold standard segmentations, may be cued by more than just the statistical recurrence of forms \citep{Todd2019,panther_morphological_2023}; we return to this point in Analysis 2 (\Cref{sec:analysis2}). 

NMS perform slightly worse than Morfessor on words formed by compounding, and notably worse on words formed by affixation. One possible interpretation of this result is that NMS are not as good at tracking statistical recurrence as Morfessor -- hence the worse performance on both categories -- but make up for this shortcoming to some extent in compounds by being sensitive to additional cues to morphological structure  \citep{panther_morphological_2023}. The fact that NMS' difficulties are concentrated in words formed by affixation suggests that they may struggle specifically with recognizing affixes as independent of stems. A finer-grained inspection suggests that this may be related to issues of affix position, allomorphy, and/or frequency: NMS perform as well as Morfessor on words containing the highly frequent causative prefix \textit{whaka-} (precision/recall for Morfessor: 0.95/0.93; for NMS: 0.95/0.93), which has no allomorphs, but perform worse on words containing passive or nominalizing suffixes (precision/recall for Morfessor: 0.90/0.89; for NMS: 0.59/0.59), which have many allomorphs, including some that are quite infrequent.

\subsection{Analysis 1B: Affix recovery}
\label{sec:analyis1-b}

To dig further into potential sources of issues with segmenting words formed by affixation, we analyze the ability of Morfessor and NMS to recover different affixes by segmenting them off. This analysis separates the causative prefix from passive and nominalizing suffixes, and subdivides passive and nominalizing allomorphs into smaller groups.

\subsubsection{Methods}
\label{sec:analysis1-b-methods}

The affixes we analyze are organized into groups based on word position, status as default/non-default allomorph, and consistency with a major phonological template. The groups also vary in frequency. We define the type frequency of an affix group as the proportion of the 19,595 words for which \citeapos{oh_non-maori-speaking_2020} MS segmented off a morph with the same form as some affix in the group, at the appropriate word edge. We similarly define the token frequency of an affix group as the proportion of tokens in the MAONZE corpus \citep{king2011maonze} and the M\={a}ori Broadcast Corpus \citep{Boyce2006} that correspond to words for which the MS separated off some affix from the group.\footnote{We follow \citet{oh_non-maori-speaking_2020} in using Simple Good-Turing smoothing \citep{Gale1995} to ensure that words from the dictionary that were not mentioned in the corpora have a non-zero token frequency.} Type frequency is relevant for Morfessor, and both type and token frequency may be relevant for NMS.

For each affix group, we measure the rate at which Morfessor and NMS successfully recover affixes in that group by segmenting them off words. We assign each word in the affixation category to one or more groups based on the affix(es) in its gold standard segmentation. For a word in a given group, a learner successfully recovers the affix pertaining to that group if their segmentation contains a boundary at the site between the affix and the rest of the word, without also containing any boundaries at sites within the affix. The segmentation of the stem is irrelevant: a learner can successfully recover an affix from a word even if their segmentation of the rest of the word does not match that represented by the gold standard segmentation. We measure the rate of affix recovery for a group as the proportion of words in the group for which the affix is successfully recovered.

\subsubsection{Results}
\label{sec:analysis1-b-results}

The affix recovery rates for each learner across the various affix groups are shown in \Cref{tab:affix-analysis}.

\begin{savenotes}
\begin{table*}
\centering
\caption{Affix recovery rates for Morfessor and NMS across different affix groups. Affix groups vary in terms of position, status of allomorphs as default/non-default, consistency of allomorphs with major phonological templates, and frequency of occurrence (proportion of types / tokens affixed by that form).}
\label{tab:affix-analysis}
\resizebox{\textwidth}{!}{
    \begin{tabular}{@{}llrrrr@{}}
    \toprule
    & & \multicolumn{2}{c}{Frequency} & \multicolumn{2}{c}{Affix recovery} \\
    \cmidrule(lr){3-4} \cmidrule(lr){5-6}
    Affix(es) & Allomorph group & \multicolumn{1}{c}{type} & token & Morf. & NMS \\
    \midrule
    \textit{whaka-} & -- & 0.142 & 0.017 & 0.983 & 0.976 \\
    \textit{-tia}, \textit{-tanga} & default, template\footnote{\label{fn:dialects}Dialects differ in terms of whether the default thematic consonant is <t>, <h>, or <ng>, though it is most commonly <t> \citep{Harlow2007}.} & 0.128 & 0.006 & 0.995 & 0.783 \\
    \textit{-hia}, \textit{-ngia}, \textit{-hanga} & default, template\footref{fn:dialects} & 0.064 & 0.005 & 0.995 & 0.688 \\
    \textit{-a}, \textit{-nga}\footnote{\label{fn:nga}\textit{-nga} is both a passive suffix and a nominalizing suffix. As a passive suffix, it is not a default allomorph, but as a nominalizing suffix, it is. Our analysis of \textit{-nga} is restricted to its occurrence as a nominalizing suffix.} & default, non-template & 0.034 & 0.011 & 0.907 & 0.293 \\
    \textit{-kia}, \textit{-mia}, \textit{-ria}, \textit{-whia}, \textit{-kanga}, \textit{-manga}, \textit{-ranga} & non-default, template & 0.017 & 0.003 & 0.702 & 0.553 \\
    \textit{-ia}, \textit{-na}, \textit{-ina}, \textit{-hina}, \textit{-kina}, \textit{-whina}, \textit{-anga} & non-default, non-template & 0.016 & 0.002 & 0.739 & 0.370 \\
    \bottomrule
    \end{tabular}
}
\end{table*}
\end{savenotes}

Both Morfessor and NMS have extremely high recovery rates for \textit{whaka-}. This is not surprising, as it is extremely frequent, in terms of both types and tokens. For NMS, it is also highly salient due to its position at the beginning of verbs that often appear utterance-initially as imperatives (e.g., \textit{whakarongo mai!} `listen!') and its appearance in place names (e.g., Whakatane) and the well known and highly culturally significant word \textit{whakapapa} `genealogy' \citep{Oh2023}. There is also reason to believe that NMS may be particularly sensitive to prefixes such as \textit{whaka-} because they have been shown to apply a bimoraic template when segmenting the first morph in a word \citep{panther_morphological_2023}.

While Morfessor and NMS have near-identical rates for the causative prefix \textit{whaka-}, their recovery rates for allomorphs of the passive and nominalizing suffixes diverge, with NMS being less successful than Morfessor. One possible reason for this divergence is that NMS may be less sensitive to suffixes than prefixes. Another possible reason may stem from NMS being sensitive to token frequency rather than just type frequency like Morfessor, as the passive and nominalizing suffixes have much lower token frequencies than type frequencies, both in absolute terms and in relation to the corresponding frequency of \textit{whaka-}. From a statistical learning perspective, a morph needs to be experienced sufficiently often in a range of environments before a learner can reliably identify and extract it, so lower experiential frequency by NMS compared to Morfessor would yield noisier segmentations. This difference is magnified by the fact that Morfessor has perfect memory of all types it has encountered at each point of the learning process, which is not the case for NMS.

Zooming into the allomorphs, for Morfessor there is a clear separation between default and non-default. This separation is driven by frequency: allomorphs in a given affix group can be recovered reliably if and only if they recur across sufficiently many types. After adjusting for frequency, there are no major differences between affix groups based on consistency with a major phonological template, nor phonological shape generally (e.g., passive CVV vs.\ nominalizing CVCV: recovery rates 0.964 and 0.966, respectively). This is not surprising: Morfessor's naive statistical learning algorithm has no access to phonological templates, and is based primarily on frequency.

For NMS on the allomorphs, there is also a relationship between affix recovery rate and frequency, but it is more gradient, reflecting differences in the experiential frequency and memory of NMS compared to Morfessor. The correlation is not perfect, however. The affix recovery rate is extremely low for the default allomorphs that are not consistent with a template, in spite of their high token frequency. It is also higher than expected for the non-default allomorphs that are consistent with a major phonological template, in comparison to those that have almost identical frequency but are not consistent with a template. These results suggest that NMS are sensitive to major phonological templates, giving them an advantage in recognizing allomorphs that are consistent with them. 

Furthermore, since the default allomorphs that are not consistent with a template are also short -- with one simply having the shape V -- the fact that NMS recover them less successfully suggests a sensitivity to phonological shape generally. That is, NMS may find morphs less salient the less phonological content they have and/or the less their syllables resemble the canonical CV shape. This suggestion is further supported by the fact that NMS are less successful at recovering passive suffixes with a CVV shape than nominalizing suffixes with a CVCV shape (rates 0.669 and 0.866, respectively).

\section{Analysis 2: Other cues}
\label{sec:analysis2}

Analysis 1 showed that NMS are sensitive to templates, both at the word level (reduplication) and at the morph level (minimality constraints; allomorphs that follow a phonological template or feature syllables with canonical CV shape). Morfessor shows no such sensitivity, as its underlying generative model does not incorporate templates, and thus underperforms when segmenting words that invoke templates in some way.

However, templates appear not to be the only reason that Morfessor underperforms. In \Cref{sec:analysis1-a-results}, we observed that Morfessor's performance on compounds was lower than might be expected, given that they follow its underlying assumption of morphological concatenativity. Based on this observation, we suggested that the morphological structure of M\={a}ori may be cued by more than the statistical recurrence of forms, consistent with previous results showing that MS segmentations are sensitive to aspects such as the presence of long vowels \citep{Todd2019, panther_morphological_2023}. 

Here, we explore this suggestion further by comparing Morfessor's performance on real M\={a}ori words, which may contain such additional cues to morphological structure, to its performance on artificially constructed pseudo-M\={a}ori words, which are governed by the same patterns of statistical recurrence of morphs but lack any additional cues to morphological structure. This analysis reveals the extent to which such additional cues exist in real M\={a}ori and the extent to which they present issues for Morfessor. In doing so, it generalizes conclusions from \Cref{sec:analyis1} that the suitability of Morfessor to a particular language -- and, by extension, the extent to which statistical learning by non-speakers of that language may be based purely on tracking of statistical recurrence -- is dependent upon the morphological structure of the language. 

\subsection{Data}
\label{sec:analysis2-data}

In this analysis, we focus entirely on words that follow Morfessor's underlying assumption of concatenativity. We do not include words that invoke templates at the word or morph level, since the analysis in \Cref{sec:analyis1-a} already established that Morfessor underperforms in the presence of such templates.

The analysis is based on the `polymoraic' group of \citet{panther_morphological_2023}, excluding words with morphs containing more than 3 syllables. This includes a total of 1,292 words, comprising 1,199 of the 1,204 compounds that we analyzed in \Cref{sec:analyis1}, as well as an additional 93 words that \citeapos{oh_non-maori-speaking_2020} MS analyzed as simplex.

For the analysis of pseudo-M\={a}ori, we generated 1,000 different sets of 1,292 words each through concatenating morphs, based on the statistical properties of the 1,292 real M\={a}ori words (see \Cref{sec:analysis2-methods}). For each set, the generative process provided us with ground-truth segmentations, which we compare to those provided by a Morfessor model trained over the set. For the analysis of real M\={a}ori, we similarly compare the gold standard MS segmentations of the 1,292 words to those provided by a Morfessor model trained on those words (as opposed to the full lexicon from \Cref{sec:analysis1-data}).

\subsection{Methods}
\label{sec:analysis2-methods}

To generate each set of pseudo-M\={a}ori words, we used the same probabilistic process as is assumed by Morfessor's underlying generative model. This process works in a bottom-up fashion across several structural levels, first concatenating phonemes into syllables, then concatenating syllables into morphs, and finally concatenating morphs into words. Types of one level are drawn with replacement from an inventory, according to an inverse power law (Zipfian) probability distribution, and concatenated to form a type of the next level. The types at each level are unique: if a proposed type already exists, a new one is generated instead.

We generated each set of pseudo-M\={a}ori words with constraints based on real M\={a}ori, in two main ways. First, we constrained the pseudo-M\={a}ori words to have the same statistical recurrence properties as real M\={a}ori, by using an inventory probability distribution at each level that was inferred from the set of real M\={a}ori words (see \Cref{appendix:powerlaw} for details). Second, we constrained the types at each level to have the same form properties as real M\={a}ori. Specifically: at the phoneme level, we used the same 10 consonants and 10 vowels as real M\={a}ori (\Cref{sec:background-maori}); at the syllable level, we only generated syllables of shape CV and V; at the morph level, we generated the same number of monosyllabic, disyllabic, and trisyllabic morph types (respectively) as there are in the real M\={a}ori set of words; and at the word level, we used each real M\={a}ori word as a template for a pseudo-M\={a}ori word, ensuring that they matched in terms of the number of morphs and the number of syllables in correspondingly-ordered morphs. 

As in \Cref{sec:analysis1-a-methods}, our analysis is based on comparing Morfessor segmentations to a gold standard. We again use macro-averaged boundary precision and recall as the metric for this comparison.

\subsection{Results}
\label{sec:analysis2-results}

\begin{figure}
\centering
\includegraphics[scale=0.48]{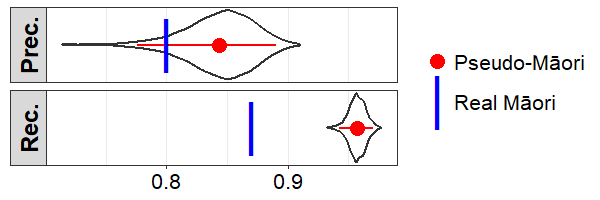}
\caption{Distributions of macro-averaged precision and recall for Morfessor's segmentations of 1000 sets of pseudo-M\={a}ori words, in comparison to its performance on corresponding words from real Māori (blue lines). Red points show mean performance on pseudo-M\={a}ori and red lines show 95\% percentile intervals.}
\label{fig:analysis2}
\end{figure} 

\Cref{fig:analysis2} shows the distributions of macro-averaged precision and recall for Morfessor's segmentations on the 1000 sets of pseudo-Māori words, together with the precision and recall for its segmentations of the corresponding real M\={a}ori words (when training is restricted just to those words). It is immediately apparent that recall is higher than precision, indicating occurrences of oversegmentation that are not balanced by undersegmentation as was the case in \Cref{sec:analysis1-a-results}. This is likely a consequence of the training set being much smaller (1,292 words as opposed to 19,595); since the same pattern is seen across real M\={a}ori and psuedo-M\={a}ori, it does not appear to reflect influences of non-statistical cues to morphological structure.  

Morfessor is better able to accurately segment pseudo-M\={a}ori than real M\={a}ori. Numerically, both precision and recall are higher for pseudo-M\={a}ori (mean precision: 0.84; recall: 0.96) than for real M\={a}ori (precision: 0.80; recall: 0.87). The advantage for pseudo-M\={a}ori is especially strong for recall, where performance on all 1,000 sets of words far exceeds that on real M\={a}ori. This strong advantage in recall is not driven by increased oversegmentation of pseudo-M\={a}ori relative to real M\={a}ori, because it is not accompanied by a concomitant disadvantage in precision; rather, it reflects the fact that boundaries in pseudo-M\={a}ori are cued by recurrence statistics, which Morfessor tracks. That is, Morfessor is best able to segment words when they come from a language that closely adheres to the statistical principles of structure that it assumes.

It follows that Morfessor's worse performance on real M\={a}ori is likely due to failure to identify boundaries that are cued by something other than morph recurrence statistics. This result therefore confirms the suggestion from \Cref{sec:analysis1-a-results} that the morphological structure of M\={a}ori may have alternative cues, though it does not indicate precisely what they may be. Past research has shown that NMS are sensitive to cues such as bimoraic templates and the presence of long vowels in the segmentation of compounds \citep{panther_morphological_2023}, and it is likely that this sensitivity explains why they were more successful at segmenting compounds than affixed words in Analysis 1A.

\section{Discussion \& conclusions}
\label{sec:discussion}

We have examined morphological segmentations of M\={a}ori by Morfessor and non-M\={a}ori-speaking New Zealanders (NMS), across words formed through a variety of morphological processes, to assess the ways in which they are affected by structural factors and the extent to which they have such effects in common. Our results show that both learners are affected by linguistic structure. In some circumstances, they are affected similarly; for example, both are successful in segmenting words formed by concatenative morphological processes (Analysis 1A), especially when highly frequent morphs are involved (Analysis 1B). In other circumstances, they are affected in opposite ways; for example, Morfessor suffers decreased segmentation performance on words that are formed via templatic processes (Analysis 1A) or that cue morphological structure by means other that statistical recurrence of forms (Analysis 2), whereas NMS see increased performance in such cases. 



These similarities and differences are important when considering the nature of human statistical learning of morphological segmentation. Since Morfessor's learning is underpinned by a set of well defined assumptions and principles (\Cref{sec:background-segmentation}), the extent to which its performance aligns with that of NMS may be taken to reflect the extent to which NMS' learning is underpinned by those same assumptions and principles. On the one hand, the similarities affirm that NMS undergo statistical learning, identifying and extracting statistically recurrent forms to build a memory-store of morphs. On the other hand, the differences show that learning for NMS does not just involve tracking statistical recurrence, but also involves inducing abstract templates about the formation of words and the shapes of (allo)morphs, as well as developing sensitivities to prominent features such as the presence of long vowels \citep{panther_morphological_2023}. 

On a practical front, these similarities and differences suggest that human statistical learning of morphological structure can be appropriately modeled by unsupervised machine learning, but only to a first approximation. When the morphological structures closely follow those assumed by the model, the morphs that the model learns can reflect the cognitive units that humans seem to operate over \citep[e.g.,][]{Virpioja2018,Lehtonen2019}. But when morphological structures vary too widely from those assumed by the model -- either within a language, based on classes of words formed by different processes, or across languages -- there is the potential for the model to miss factors that are salient to humans but that it is not equipped to handle. This is especially important as different models have different underlying assumptions, which can respond differently to variation in morphological structure \citep{loukatou_does_2022}.

The differences in the segmentation performances of Morfessor and NMS across words of different morphological structures not only inform the use of unsupervised morphological segmentation models as cognitive models, but also highlight potential factors that could be incorporated into segmentation models to improve their results. For example, inspired by the observation that reduplication templates are salient to humans but not to Morfessor, \citet{todd_unsupervised_2022} show that adding reduplication templates to Morfessor improves its ability to find reduplication in M\={a}ori words. 
Similarly, future research that dissects NMS' underlying learning mechanisms could reveal additional generalizable factors that help improve the cross-linguistic applicability of unsupervised models.
\bibliography{acl2023.bib}
\bibliographystyle{acl_natbib}

\appendix

\section{Generating pseudo-M\={a}ori: Details}
\label{appendix:powerlaw}

This appendix describes the process through which we inferred statistical recurrence properties of M\={a}ori, to use in the generation of pseudo-M\={a}ori.

We derived inventories at each level -- unique phonemes, syllables, morphs, and words -- from the segmentations provided by \citeapos{oh_non-maori-speaking_2020} fluent M\={a}ori speaker. To get the frequency distribution over types at one level, we counted occurrences within unique types at the next level. That is, we counted the number of unique syllables that each phoneme occurred in; the number of unique morphs that each syllable occurred in; and the number of unique words that each morph occurred in. We sorted each distribution by count, to obtain rank and frequency for each type, and fit an inverse power law $f(x) = ab^{-x}$ to predict frequency from rank, using nonlinear least squares. 

To sample in the generative process, we sorted the types in random order and treated those orders as ranks, overlaying the frequency from the inverse power law and then normalizing to obtain a probability distribution.

\end{document}